# Telling Human and Machine Handwriting Apart

Luis A. Leiva, Moises Diaz, Nuwan T. Attygalle, Miguel A. Ferrer,
and Réjean Plamondon, *Life Fellow, IEEE*

*Abstract*—Handwriting movements can be leveraged as a unique form of behavioral biometrics, to verify whether a real user is operating a device or application. This task can be framed as a "reverse Turing test" in which a computer has to detect if an input instance has been generated by a human or artificially. To tackle this task, we study ten public datasets of handwritten symbols (isolated characters, digits, gestures, pointing traces, and signatures) that are artificially reproduced using seven different synthesizers, including, among others, the Kinematic Theory ($\Sigma h$ model), generative adversarial networks, Transformers, and Diffusion models. We train a shallow recurrent neural network that achieves excellent performance (98.3% Area Under the ROC Curve (AUC) score and 1.4% equal error rate on average across all synthesizers and datasets) using nonfeaturized trajectory data as input. In few-shot settings, we show that our classifier achieves such an excellent performance when trained on just 10% of the data, as evaluated on the remaining 90% of the data as a test set. We further challenge our classifier in out-of-domain settings, and observe very competitive results as well. Our work has implications for computerized systems that need to verify human presence, and adds an additional layer of security to keep attackers at bay.

## I. Introduction

ONLINE fraud often involves identity theft, and most of today's security measures are weak or can be spoofed. A plausible next level of security is to identify people using behavioral information, since it is much harder to copy or imitate. However, recent work [1] noted that synthetic mouse movements can be used effectively as reply attacks. This kind of machine impersonation is extremely scalable because in a matter of seconds it is possible to generate hundreds of thousands of artificial data samples (or *specimens*) for

This work was supported in part by the Pathfinder Program of the European Innovation Council under Grant 101071147; in part by the PID2023-146620OB-I00 Project funded by MICIU/AEI under Grant 10.13039/501100011033; in part by the European Union's ERDF Program; and in part by the CajaCanaria and LaCaixa under Grant 2023DIG05. This article was recommended by Associate Editor P. Chattopadhyay. *(Corresponding authors: Luis A. Leiva; Moises Diaz.)*
Luis A. Leiva is with the Computer Science Department, University of Luxembourg, 4365 Esch-sur-Alzette, Luxembourg (e-mail: luis.leiva@uni.lu).
Moises Diaz and Miguel A. Ferrer are with the Instituto Universitario para el Desarrollo Tecnológico y la Innovación en Comunicaciones, Universidad de Las Palmas de Gran Canaria, 35017 Las Palmas de Gran Canaria, Spain (e-mail: moises.diaz@ulpgc.es).
Nuwan T. Attygalle is with the Louvain Research Institute in Management and Organizations, Université catholique de Louvain, 1348 Louvain-la-Neuve, Belgium.
Réjean Plamondon is with the Department of Electrical Engineering, Polytechnique Montréal, Montréal, QC H3T 0A3, Canada.

malicious use. Furthermore, some websites are starting to ask their users to solve some form of handwritten captcha [2], [3], based on the assumption that handwriting input is a very natural action for humans.

In this context, we can think of a new form of biometric authentication for online services based on handwritten symbols, such as gestures (geometric shapes or marks), characters, digits, or signatures that users would have to enter on some touch-sensitive surface, such as a mobile phone or tablet. The main advantage is that these symbols are short and easy to enter for humans. In addition, gestures and digits are language-independent, so they are equally easy to learn for everyone. Finally, signatures comprise overlearned ballistic movements that people articulate almost without thinking. Therefore, we can expect an increasing adoption of some form of handwriting-based verification or reauthentication mechanisms in the future.

Several computational models have been developed to create synthetic on-line handwriting.[1] For example, generative adversarial networks (GANs) [1], cognitive models [4], and computational models based on neuromotor principles [5] have allowed for a sophisticated and human-like generation. In this article, as described later, we consider several state-of-the-art (SoTA) handwriting generation models, including, e.g., GANs, Transformers, and Diffusion models.

In previous work we compared different model architectures to distinguish between human and machine-generated handwriting movements [6], including convolutional neural networks (CNNs) and recurrent neural networks (RNNs). We proposed a shallow RNN architecture that used the pen-tip velocity as sole input feature, motivated by the fact that velocity is a significant feature in handwriting recognition [7], considered the fundamental motor control variable in human handwriting [8]. That classifier achieved 96% Area Under the ROC Curve (AUC) (averaged over 4 datasets). In this article, we propose a classifier based on that RNN architecture but using *unprocessed* (nonfeaturized) movement sequences as input (Fig. 1). Doing so allows the model to learn the best internal representation, instead of doing the feature engineering ourselves. We substantially challenge the robustness of this classifier using more, larger, and more diverse public datasets, and using new synthetic data generation approaches. We were able to improve model performance systematically, with over 98% AUC (averaged over 10 datasets and 7 different synthesizers). Importantly, our findings are consistent across a large variety of on-line handwritten symbols articulated on different devices (e.g., smartphone and tablet) using different input methods (e.g., stylus and finger) and across several SoTA data generation methods, as explained later. We also conducted few-shot and out-of-domain classification experiments to prove further the generalization capabilities of our classifier. In sum, this article makes the following contributions.

---

[1]The handwriting recognition community uses the term "on-line" to denote sequential data, as opposed to the term "off-line" which denotes nonsequential data, such as a scanned image of a handwritten signature.

1) We propose a classification model for conducting reverse Turing tests that takes as input just plain trajectory data.
2) We provide comprehensive experiments that demonstrate the robustness and generalizability of the model, including few-shot and out-of-domain scenarios, and analyzes of movement complexity.
3) Our results are consistent across ten datasets and seven handwriting generators.

Ultimately, this research has key implications for researchers interested in understanding the differences between human and machine movements. If online services are to rely on behavioral data to prevent fraud, then we have to ensure that it is possible to distinguish between human and synthetic data reliably. This idea is similar in soul to the concept of "liveness detection" in the biometrics community [9]. In short, biometric systems without liveness detection may be fooled very easily and the consequences might be fatal [10]. Our classifier achieves a remarkably high performance over all the datasets and all the generation approaches considered, therefore it can be seen as a compelling production-ready reverse Turing test, where a computer (instead of a human) distinguishes between human and machine-generated data.

## II. Related Work

Among soft biometrics technologies, the ones based on handwriting movements are increasingly being used to authenticate people. Signatures are perhaps the most popular approach, however there are more options to verify online users by means of handwriting input. For example, Acien et al. [1] considered mouse movements for online verification coupled with a captcha. In Highlighting CAPTCHA [2] the user had to trace an obfuscated word with a stylus, and μcaptcha [3] required the user to trace a math equation.

To verify legitimate human presence on websites and online services, a simpler and more fundamental approach is the one we envision in this article: Just ask the user to handwrite anything and verify if the resulting data was actually produced by a human. Note that the goal is to allow a computer to recognize *any* human handwriting data from *any* machine-generated data, i.e., to develop a reverse Turing test. A key aspect to consider is that handwritten data is of sequential nature, therefore we can assume they will be spatiotemporal sequences of *(x, y, t)* tuples. This combination of spatial and temporal information is hard to replicate. This relevant aspect was highlighted by Elarian et al. [11], who concluded that on-line handwriting velocity is particularly difficult to simulate reliably.

One of the most successful techniques to generate realistic handwriting data are movement simulation approaches based on human neuromotor control and feedforward models of locomotion. A notable approach in this regard is the Kinematic Theory [12] and its associated Sigma–Lognormal *(ΣΛ)* model. According to this theory, *aimed* human movements (i.e., "movements with intent") are defined by elementary movement units, also known as *components* or *primitives*, that are superimposed to produce the resulting trajectory. The *ΣΛ* model has been successfully used for synthesizing handwriting gestures [5] and signatures [13], among other types of human movements.

The majority of previous work in handwriting generation has focused on Latin scripts, although we can also find language-specific synthesizers for Bengali and Devanagari [14], Chinese [15], and Arabic [16]. Autoregressive methods for on-line handwriting generation have become popular, since they allow end-to-end modeling based on examples, following machine learning principles.

For example, some researchers have proposed RNNs in combination with Gaussian mixture models to generate plausible handwritten texts [17], [18]. Other researchers have proposed variational autoencoders (VAEs) to generate sketches [19] and diagrams [20] in addition to handwritten text [21]. Disentangling writer and character styles has also been the subject of recent research [15], [22], [23]. More recently, researchers have investigated Transformers [24], [25] and Diffusion models [26], [27], achieving compelling results. We showed in previous work [6] that on-line handwriting can be classified more confidently than off-line handwriting, and proposed a recurrent model architecture that was tested on four datasets. As discussed in Section I, that model required a feature engineering step prior to model training, as it relied on the pen-tip velocity as input data. In this work, we use the same model architecture but use plain trajectory data as input, thereby allowing the model to learn the best internal representation. We also analyze six additional datasets that include more challenging and more diverse types of handwriting data, such as mouse and mid-air movements. We also consider more ways of synthesizing handwriting data in addition to the *ΣΛ* model: analytical *ΣΛ* reconstructions, affine transformations, GANs, Transformers, and Diffusion models. Finally, we also conduct few-shot and out-of-domain experiments that, together, demonstrate the robustness and generalizability of our model. In sum, this article substantially advances our previous work and sets a new state-of-the-art in the task of telling human and machine handwriting apart. Put simply, our work is essential to prevent spoofing attacks, when a malicious attacker tries to defeat a biometric system through the introduction of fake biometric samples.

## III. Methodology

We first introduce the different synthetic handwriting generation approaches considered, and then we describe our classifier to tell human and computer handwriting apart.

### A. Handwriting Synthesizers

We use three different approaches to synthetically reproduce human handwriting: two neuromuscular-based methods, four SoTA data-driven generation methods as reported by Diaz et al. [28], and an affine-based synthesis method. We produced one synthetic version for each human sample in the datasets using each synthesizer.

*1) Neuromuscular-Based Synthesis:* We consider two approaches based on the $\Sigma\Lambda$ model, which has a biological background and, as mentioned before, has been used extensively for the generation of many kind of handwriting movements.

*Kinematic-wise reconstruction* $\hat{r}$ is the analytical reconstruction of a given real sample $r$, according to the extracted $\Sigma\Lambda$ components.[2] This reconstruction has the same duration as the input trajectory, but it is resampled uniformly at the same input frequency. The pen-tip velocity is analytically parameterized using a lognormal extractor from ScriptStudio [29], an application that uses the "Robust XZERO" algorithm. Since a complete treatment of this approach is outside of the scope of this article, we refer the interested reader to the work of O'Reilly and Plamondon [29] for more details.

*Kinematic-wise synthesis* $s_k$ takes as input an analytical reconstruction and applies random perturbations to its $\Sigma\Lambda$ components, according to the expected bounds of human variability. Several ranges are reported in previous work to generate synthetic samples [5], [13], [30], among which we followed the ones proposed in [13] as they have proven to be the more general ones, allowing us to generate realistic characters, digits, gestures, pointing traces, and signatures. Then, each component's duration is randomly expanded or contracted by 10% of its original duration at the same input frequency. To introduce further variability, each component is displaced up to a maximum distance of 10% of its previous position.

For both cases, i.e., $\hat{r}$ and $s_k$, all components are concatenated in order to produce the final synthetic trajectory. Fig. 2 depicts some handwriting examples both in the spatial (left columns) and temporal (right columns) domains.

*2) Data-Driven Synthesis:* We describe the four SoTA models we considered, followed by the specific training details, as indicated in their respective papers. All models were trained using Adam optimizer with decay rates $\beta_1 = 0.9$, $\beta_2 = 0.999$, unless stated otherwise below. None of the models have been tested before on our handwriting datasets (Table I).

*BeCAPTCHA* [1] $s_b$ is a GAN designed to distinguish between mouse movements generated by human and bots solving Google's visual captchas. The GAN takes as input a seed vector of 100 random numbers and produces a sequence of $\{x, y\}$ coordinates assuming a fixed sampling rate of 200 Hz [1]. Under the hood, BeCAPTCHA uses the $\Sigma\Lambda$ model to extract features from mouse dynamics. We trained BeCAPTCHA for 50 epochs using a batch size of 128 samples and learning rate $y = 0.0002$.

*DeepWriteSYN* [21] $s_w$ is another SoTA handwriting synthesizer. It comprises two modules: 1) temporal segmentation of handwriting data into short-time segments, using the $\Sigma\Lambda$ model; and 2) on-line synthesis of those short-time segments using Sketch-RNN [19], a sequence-to-sequence VAE. The main advantage of DeepWriteSYN is that it operates in short-time segments, overcoming the length limitation and drifting issues of Sketch-RNN [31]. We trained DeepWriteSYN for 100 epochs using a batch size of 100 samples and learning rate $y = 0.0001$. The loss function is a combination of a reconstruction term and the (weighted) Kullback–Leibler divergence ($w_{KL} = 0.25$). For generation, we used a temperature $\tau = 0$ (deterministic output), as suggested by Tolosana et al. [21].

*Disentangled Transformer* (SDT) [25] $s_t$ is a recent synthesizer that consists of a dual-head style encoder, based on ResNet18 [32] and a Transformer content encoder, and a Transformer decoder with 15 attention heads. It employs two complementary contrastive objectives to extract the style commonalities of reference handwriting samples, which allow it to generate realistic synthetic samples. We trained SDT for 200000 steps (which corresponds between 400 and 4000 epochs, depending on dataset size) using a batch size of 32 samples and learning rate of $y = 0.0002$.

Finally, *Diffusion model for handwriting generation* (DHG) [26] $s_d$ is another SoTA data-driven synthesizer that we considered for our study. The model consists of two parts: 1) a text and style encoder based on MobileNetV2 [33], to represent the desired text and stylistic handwriting features; and 2) a diffusion probabilistic model comprising a UNet-based backbone [34]. We trained DHG for 60 000 steps (which corresponds between 400 and 4000 epochs, depending on dataset size) using a batch size of 96 samples and learning rate of $y = 0.0001$.

*3) Affine Synthesis:* We also consider a basic method based on affine distortions for on-line data. Affine synthesis distorts the input trajectory by applying a cognitive model [4] followed by naïve transformations. Like in the previous method, a handwriting trajectory is first divided into components. Then, a sinusoidal transformation is applied to each component, which introduces intrapersonal variability. Next, the spatial coordinates are interpolated by a cubic spline and resampled at 100 Hz, which randomly lengthens or shortens the trajectory up to 10% of their original duration. Then, intercomponent variability is added by randomly displacing each component. Finally, the components are merged, and two affine geometric transformations are applied: one mimics handwriting slant and other mimics handwriting skew, producing the resulting artificial movement data $s_a$.

### B. Classifier Architecture

Since our goal is a binary classification task (i.e., telling human and computer handwriting apart), instead of classic handwriting classification tasks (e.g., decoding symbol classes or identifying the writers), we considered the architecture proposed in our previous work [6] but considering trajectory-based representations in addition to velocity-based (Section IV-B).

Our classifier has a single gated recurrent unit (GRU) hidden layer [35] with 100 neurons (embedding size) followed by a Dropout layer with a drop rate $q = 0.25$, for regularization purposes, and followed by a single output neuron, since the classifier outputs the probability of a movement being

---
[2] A handwriting movement is decomposed into several submovements, or components, each represented by a lognormal function that has two peripheral and four control parameters.

4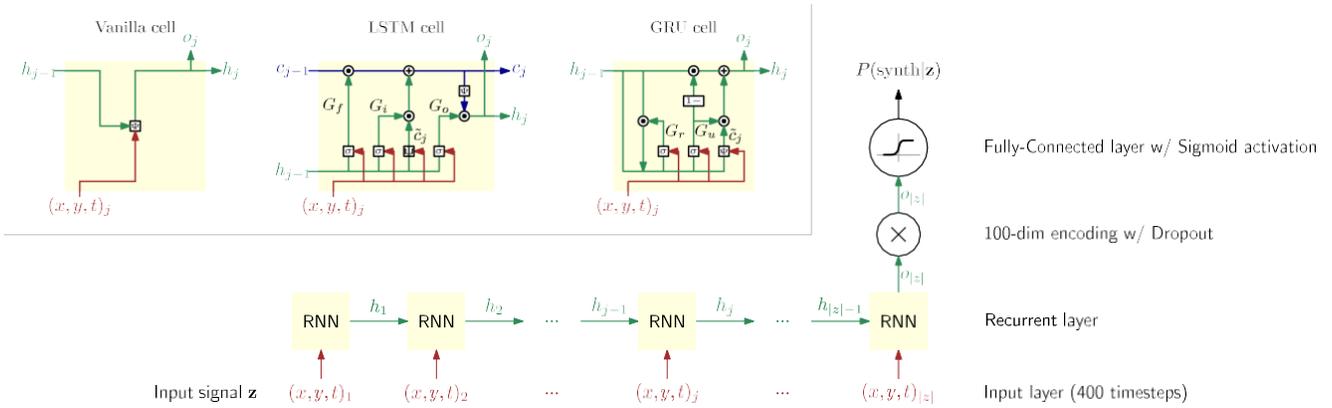

Fig. 1. Overview of RNN cells (top left) and our model architecture for sequence classification.

synthetic. Specifically, a movement **z** is classified as being synthetic when $P(\text{synth}|\mathbf{z}) > 0.5$ and human otherwise. Fig. 1 depicts the architecture. Essentially, GRU is a simplified version of the popular long short-term memory (LSTM) cell [36]. In the following, we briefly explain the main differences between these RNN cells.

Vanilla RNN cells consist of a cyclic (or *loop*) connection, which enables the model to update the current state based on past states and current input data. However, vanilla RNNs are not capable of handling long-term dependencies. To solve this issue, the LSTM and GRU cells were developed. LSTM cells have two kinds of hidden states [37]: a "slow" state $c_j$ that keeps long-term memory, and a "fast" state $h_j$ that makes decisions over short periods of time. In practice, the activation functions *W* and *σ* are hyperbolic tangent and sigmoid, respectively. However, other nonlinear functions have been promoted in the research literature; see *A note on activation functions* later for a brief discussion. On the other hand, GRU cells control the flow of information like LSTM cells, but without having to use memory, since the full state vector is output at every time step. Furthermore, performance-wise the GRU cell is on par with the LSTM in many problems [38] but it is computationally more efficient because of its less complex structure. Therefore, our classifier uses GRU cells for its recurrent layer.

*Note on Activation Functions:* Researchers have favored the rectified linear unit (ReLU) as nonlinear activation function since it considerably speeds up training convergence. In our experiments, however, we noted that ReLU does not achieve good performance for sequence classification tasks. Others have also noted that ReLU is not a good activation function for RNNs [39], mainly because of the vanishing gradient and exploding gradient problems [40], [41]. To overcome these problems, we need an activation function whose second derivative can sustain for a long range and that can saturate for positive and negative gradients. Therefore, our classifier uses tanh as the main activation function and sigmoid activation for the recurrent steps.

## IV. EXPERIMENTS

### A. Datasets

We consider ten large-scale public datasets of isolated symbols, digits, and characters as well as pointing movements and complex signatures acquired with different sensors and often at nonuniform sampling frequencies. While many other datasets are also available [42], [43], [44], our selection covers a wide range of different handwriting application domains. Table I summarizes the datasets. Aiming at investigating potential effects of the input device in classification performance, we split the two device-dependent datasets $N-MMG$_S$ and BioSign for analysis as follows: $N-MMG$_S$ (stylus), $N-MMG$_F$ (finger), BioSign$_S$ (stylus), and BioSign$_F$ (finger).

All datasets comprise spatiotemporal sequences of $(x, y, t)_j$ tuples, where $(x, y)_j$ refer to the $j$th spatial coordinate of the pen tip and $t_j$ refers to the time at which the coordinate occurred, which can be provided in many different forms, for example: Unix timestamps, millisecond-precision offsets, or systematically constant values (if coordinates were acquired at a uniform sampling rate).

### B. Input Representations

We consider two data representations as input to our classifier. The first representation is pen-tip "velocity," an engineered feature proposed in previous work [6], defined as the Euclidean distance between two consecutive points divided by their time difference. The second representation is pen-tip "trajectory," computed as the offsets between consecutive points: $(\Delta x, \Delta y, \Delta t)$, to make the classifier independent of the scale of the input device. Notice that, by feeding the classifier with nonfeaturized data, just the differences between consecutive spatiotemporal points, it has to learn the best internal (hidden) representation of the input sequence for classification.

### C. Classifier Training

Our proposed model is trained with the Adam optimizer using decay rates $\beta_1 = 0.9$, $\beta_2 = 0.999$ and learning rate $\gamma = 0.0005$. The loss function to minimize is binary cross-entropy, since our task is a two-class classification problem. We feed the classifier in batches of 128 sequences each and use up to 400 epochs for training. We also use early stopping with patience of 40 epochs, to prevent overfitting, with classification accuracy as the monitoring metric. We train the classifier on 70% of the data, validate on 10% of the data, and test on the remaining 20% of the data, which simulate unseen data. These splits are random, aiming for a user-independent classifier, which is a more challenging and a more general scenario than a user-dependent classifier. The splits are also stratified, to ensure that the ratio in the training/validation/testing partitions is always 50% human and 50% synthetic samples.

TABLE I
SUMMARY OF THE EVALUATED DATASETS. THE "TRIALS" COLUMN DENOTES THE NUMBER OF EXECUTIONS PER SYMBOL CLASS, WHEREAS "INSTANCES" DENOTES THE TOTAL NUMBER OF SAMPLES IN THE DATASET

| Dataset | Source | Movement type | Device type | Users | Classes | Trials | Instances | Rate (Hz) |
|---|---|---|---|---|---|---|---|---|
| $1-GDS | [45] | gestures | pocket PC | 11 | 16 | 30 | 5,280 | 60* |
| $N-MMG | [46] | gestures | tablet PC | 20 | 16 | 30 | 9,600 | 60* |
| Chars74k | [47] | handwriting | tablet PC | 55 | 62 | 1 | 3,410 | 100 |
| SUSIGv | [48] | signatures | ePad-ink tablet | 94 | 1 | 20 | 1,880 | 100* |
| Biosecure | [49] | signatures | digitizer tablet | 132 | 1 | 16 | 2,112 | 100 |
| MCYT-100 | [50] | signatures | digitizer tablet | 100 | 1 | 25 | 2,500 | 100 |
| MobileTouch | [51] | handwriting | smartphone | 217 | 72 | 1 | 15,624 | N/A |
| BioSign | [52] | signatures | notebook and digitizer | 65 | 1 | 32 | 2,080 | 200* |
| Mouse | [53] | pointing | computer mouse | 20 | 1 | 48 | 960 | 60* |
| ProjectedSign | [54] | signatures | optical hand tracking | 80 | 1 | 20 | 1600 | 60 |

\* denotes non-uniform sampling rates.

We set a maximum capacity for our RNN to be 400 timesteps, so that longer sequences are truncated to 400 timesteps and shorter sequences are padded with zeros up to 400 timesteps. We train a model per dataset and only once, given that, according to previous experiments with different network initializations with different random seeds, we have observed no differences in classification performance.

### D. Evaluation Metrics

We report AUC and equal error rate (EER) as our main performance metrics. AUC informs about the discriminative power of any classifier, whereas EER is the rate at which a classifier is equally likely to wrongly accept a synthetic sample as human (false acceptance rate) as it is to wrongly reject a human sample as synthetic (false rejection rate).

## V. RESULTS

Our goal is to develop a reverse Turing test as general as possible, therefore we train user-independent and device-independent classifiers on all datasets. We also conduct few-shot and out-of-domain classification experiments to prove further the generalization capabilities of our classifier.

### A. Detecting Synthetic Movements

As can be observed in Figs. 3 and 4, $\Sigma\Lambda$ model reconstructions are a bit more challenging to distinguish from genuine human movements when using the pen-tip velocity as input to the classifier (see "Velocity" group). This is because the $\Sigma\Lambda$ model creates a high-fidelity representation of velocity profiles that often results in an accurate spatiotemporal trajectory. Nevertheless, we can see that when our classifier is fed with plain trajectory data, the human and artificial samples are very easy to tell apart (see "Trajectory" group in Fig. 3). With only three exceptions (SDT in Chars74k and both $\Sigma\Lambda$ reconstruction and DHG in Biosecure) where our classifier achieved 71% and 86% AUC, in the other 70 experiments it turned out that AUC was 100% or close to 100%.

On the other hand, the affine method turned out to be the easiest to detect. This is explained by the fact that the simulated timestamps do not follow plausible spatial locations, and the classifiers exploited that to their own benefit [55]. BeCAPTCHA's generated samples were also very easy to detect. Overall, we argue that it is easier to simulate a univariate time series (i.e., velocity) than a multivariate time series, i.e., sequences of (x, y, t) tuples, and as such the human-likeness aspect is more challenging to achieve in the latter case with any computerized model simulation.

### B. Few-Shot Classification

Few-shot classification aims to train a competitive recognizer using limited training data. To further challenge our classifier, we systematically train it on small stratified splits (10% and 50%) of each dataset and test it over the remaining data (90% and 50%, respectively). The results of these experiments are shown in Figs. 5 and 6. In all cases, we use a random split of 20% of the training partition as validation data. That is, for the 10% training split case, we actually use 8% for training and 2% for validation. Similarly, for the 50% training split case, we actually use 40% for training and 10% for validation.

Notice that the smaller the training split, the more likely that user-dependent samples will be left out. While this setup does not guarantee a proper user-dependent scenario, nor is it the goal of our work, it lowers significantly the chance of observing samples from a training user in the test set, considering the sample size of the training splits. As can be observed, classification performance remains about the same in all splits. This puts forward the fact that our classifier can learn an efficient decision boundary with as few training samples as 10% of the available data. Note that the test partition in these experiments is different from the one used in the previous set of experiments.

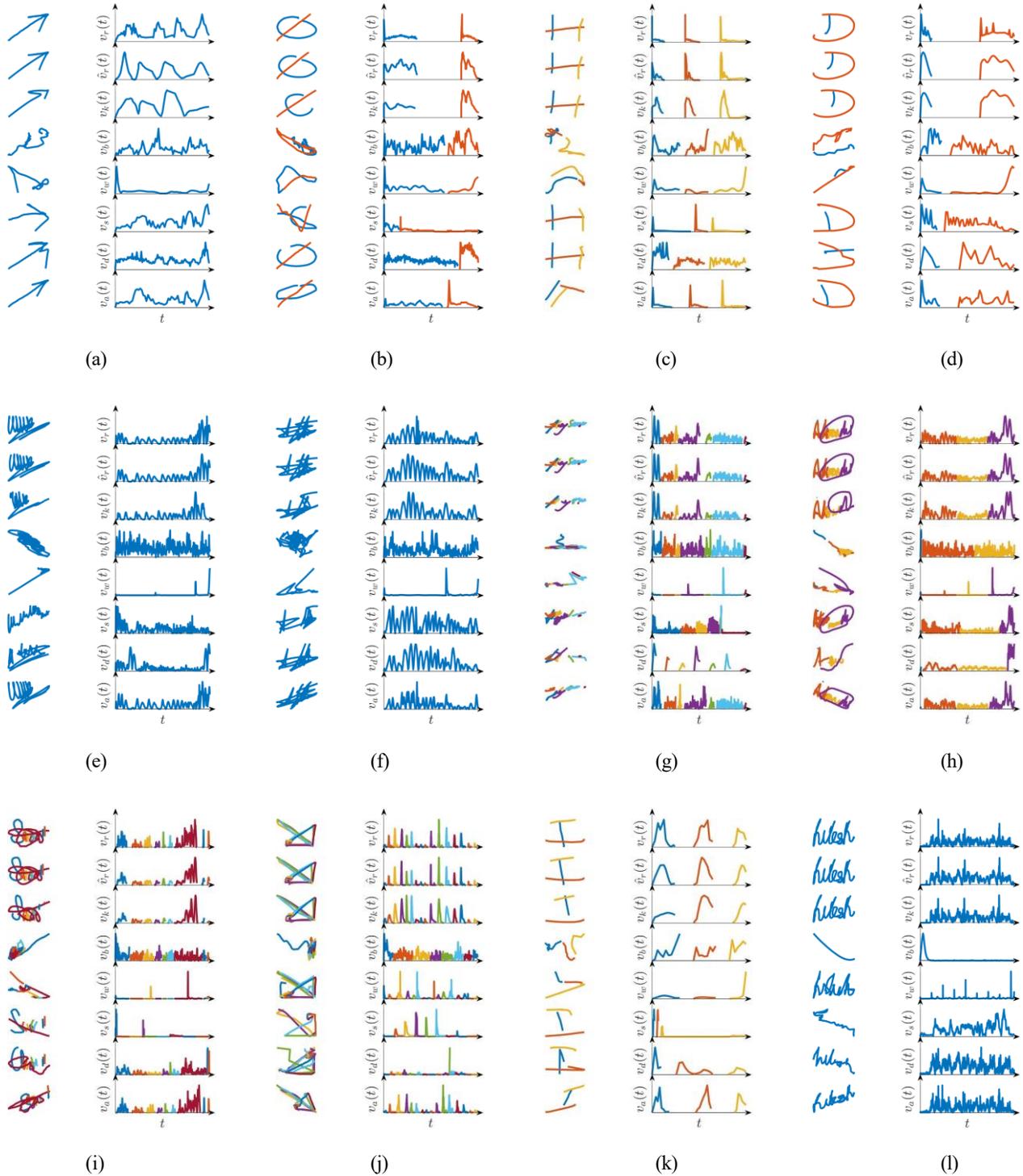

Fig. 2. Examples of movement trajectories and their corresponding velocity profiles for a real sample (top row, $v_r$) and seven synthetic counterparts, each generated with one of the proposed methods, from top to bottom: $\Sigma\Lambda$ reconstruction ($\hat{v_r}$), $\Sigma\Lambda$ synthesis ($v_k$), BeCAPTCHA ($v_b$), DeepWriteSYN ($v_w$), SDT ($v_s$), DHG ($v_d$), and affine transformations ($v_a$). (a) $1-GDS. (b) $N-MMG$_S$. (c) $N-MMG$_F$. (d) Chars74k. (e) Biosecure. (f) MCYT-100. (g) SUSIGv. (h) BioSign$_S$. (i) BioSign$_F$. (j) Mouse. (k) MobileTouch. (l) ProjectedSign.

Again, it is the nonfeaturized input representation the one that provides better performance results in all cases. The only exceptions are Biosecure, MCYT-100, and ProjectedSign at 10% for $\Sigma\Lambda$ related samples, where the velocity representation achieves better AUC results, sometimes by a large margin (see, e.g., the Reconstruction results for Biosecure and ProjectedSign at 10%). Then, as more training data are considered, the nonfeaturized representation outperforms the velocity representation in every case, often by a large margin (see, e.g.,

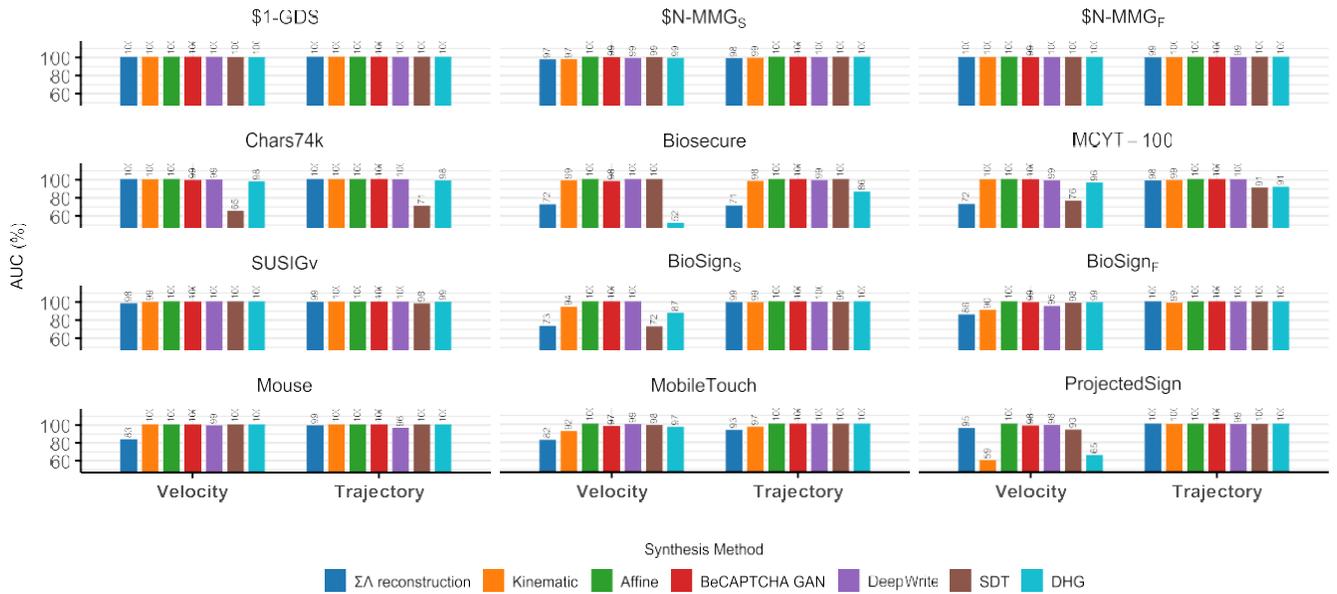

Fig. 3. Classification of human versus synthetic samples: AUC results (higher is better).

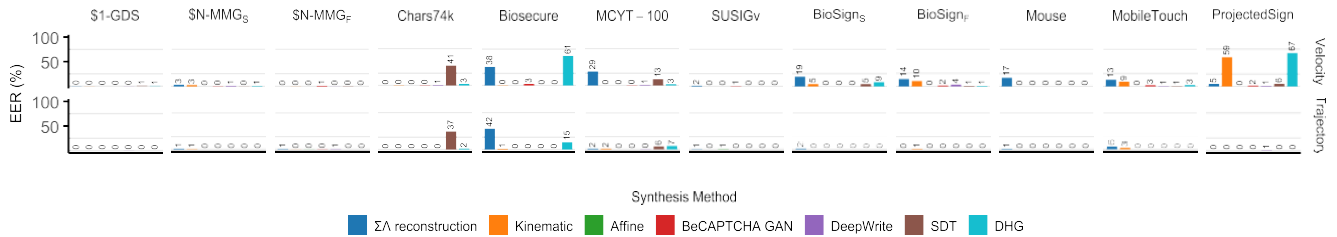

Fig. 4. Classification of human versus synthetic samples: EER results (lower is better).

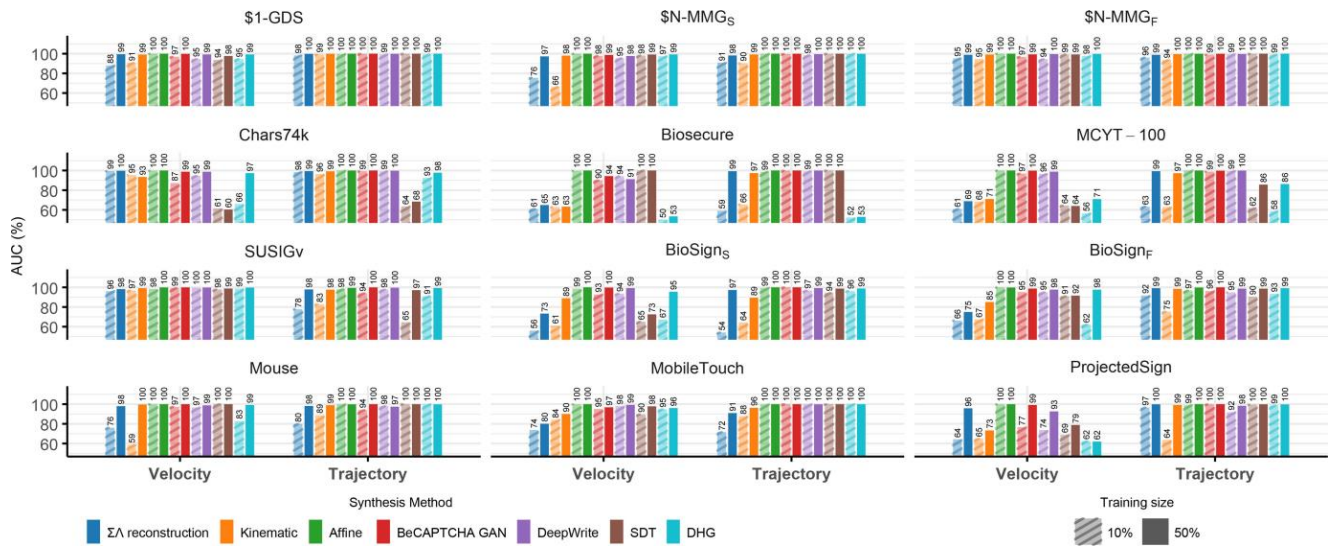

Fig. 5. Few-shot classification of human versus synthetic samples: AUC results (higher is better). The "Training size" legend group denotes the number of training instances, as a ratio of the original dataset size.

the boost in Reconstruction classification for ProjectedSign at 50%).

### C. Out-of-Domain Classification

In light of the exceptional results observed in the previous experiments, we decided to challenge our classifier further.

In out-of-domain (OOD) classification, the distribution of the test data is completely unknown to the recognizer. In contrast with the few-shot classification scenario, in OOD classification the decision boundary is potentially infinite [56] and so the training data might not efficiently reflect the test data. This scenario aligns better with our end goal of developing a general reverse Turing test, where the provenance of a handwriting movement is unknown to the classifier. We mimic this OOD scenario by testing our already trained classifiers from Section V against all the other datasets altogether. Each of the test datasets contributes with a random split of 30% of their samples. The results of these experiments are reported in Figs. 7 and 8.

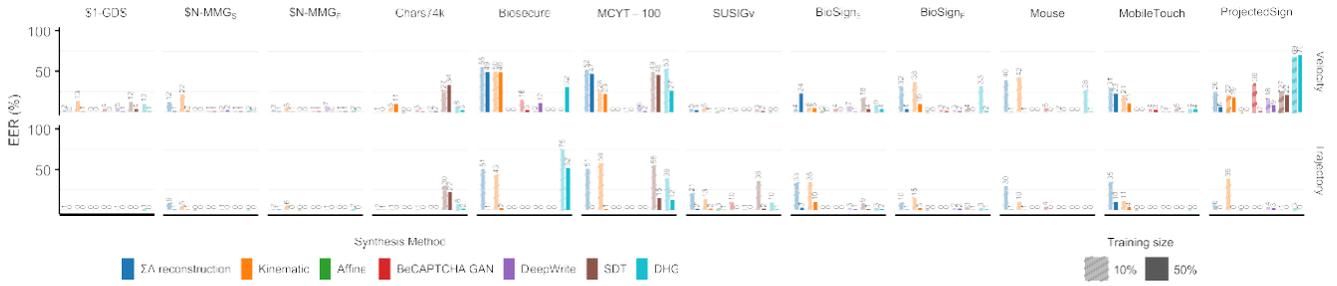

Fig. 6. Few-shot classification of human versus synthetic samples: EER results (lower is better). The "Training size" legend group denotes the number of training instances, as a ratio of the original dataset size.

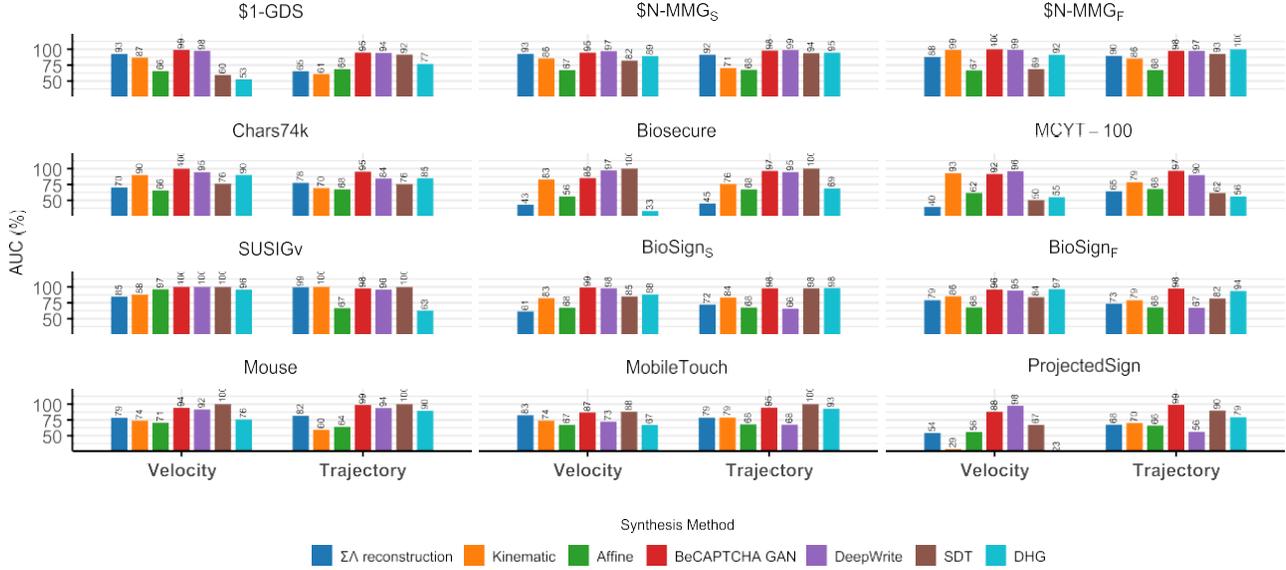

Fig. 7. OOD classification of human versus synthetic samples: AUC results (higher is better). Our classifier is trained on a single dataset, denoted in the "Source" column, then tested against a random sample from all the other datasets altogether (each dataset contributes with a random 30% of their data).

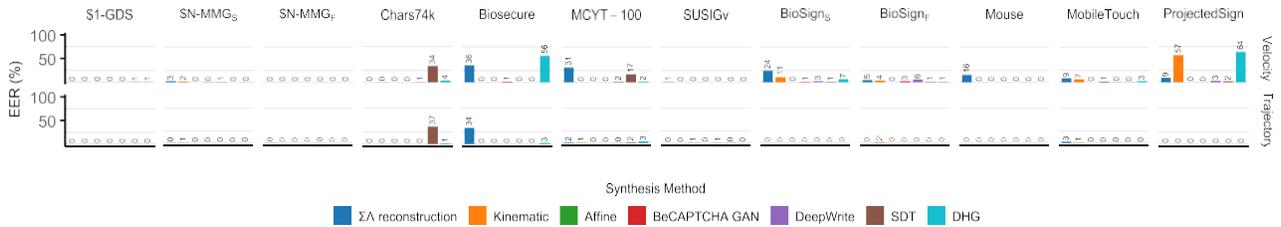

Fig. 8. OOD classification of human versus synthetic samples: EER results (lower is better). Our classifier is trained on a single dataset, denoted in the "Source" column, then tested against a random sample from all the other datasets altogether (each dataset contributes with a random 30% of their data).

We can see that all classifiers generalize reasonably well to unseen data from any of the other datasets. However, it is the classifier trained on the MobileTouch dataset the one that achieves better performance, both in terms of AUC and EER. We argue this is so because the MobileTouch dataset is very diverse in terms of symbols (72 different classes) and input devices (94 different smartphones). Also importantly, this dataset has many more training samples than the other datasets; e.g., it is three times larger than $1-GDS and almost one order of magnitude larger than ProjectedSign.

Based on our results, we would recommend training on MobileTouch data to learn a general representation of what constitutes a competent detector of synthetic handwriting movements of potentially any type, including gestures, letters, symbols, signatures, and mouse movements. Of course, if we can afford training on more datasets, this detection task would become even more accurate. By way of example, Table II shows some OOD classification examples, picked at random, of both successful and failed cases. We also report in the table the classification probability, to give a sense of how confident our classifier was in each case. It can be observed that, even when the classifier is wrong (Type I or II errors), it is so by a small margin. We remind the reader that our classifier outputs the probability of a sample **z** being synthetic, and uses $P($synth$|$**z**$) > 0.5$ for classification.

TABLE II
OOD CLASSIFICATION EXAMPLES

| Specimen (z) | Outcome | True label | Prediction | $P($synth$|$z$)$ |
|---|---|---|---|---|
| 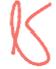 | Success | Synthetic | Synthetic | 0.9999 |
| 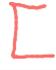 | Success | Human | Human | 0.0003 |
| 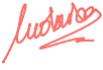 | Type I error | Human | Synthetic | 0.6989 |
| 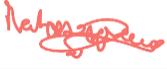 | Type II error | Synthetic | Human | 0.4305 |

TABLE III
PERFORMANCE EVALUATION RESULTS (IN %) ON ALL DATASETS COMBINED. ARROWS INDICATE THE DIRECTION OF BETTER PERFORMANCE

| Representation | AUC↑ | EER↓ | Bal. Accuracy↑ | F-score↑ |
|---|---|---|---|---|
| Velocity | 91.71 | 14.53 | 96.14 | 96.12 |
| Trajectory | 92.54 | 14.03 | 97.19 | 97.14 |

TABLE IV
PERFORMANCE EVALUATION RESULTS (IN %) ON ALL DATASETS COMBINED AND ENSURING THE SAME NUMBER OF HUMAN AND SYNTHETIC SAMPLES FOR TRAINING/VALIDATION/TESTING

| Representation | AUC↑ | EER↓ | Bal. Accuracy↑ | F-score↑ |
|---|---|---|---|---|
| Velocity | 92.13 | 5.01 | 92.13 | 92.27 |
| Trajectory | 96.03 | 2.66 | 96.03 | 96.06 |

*D. Putting It All Together*

As final experiment, we investigated further the generalization abilities of our classifier by training and evaluating in all datasets combined. In total, we used 267K samples for training, 53K for validation, and 29K for testing.

Table III summarizes the results. Again, the trajectory-based classifier outperformed the velocity-based one, although this time the differences between both approaches were not as large as in the individual experiments. We attribute this result to the significantly large scale of these experiments, more than two orders of magnitude as compared to the individual experiments. In addition, and more critically, it is important to mention that in these experiments we considered all generated samples from each synthesizer. Therefore, the proportion between human and synthetic samples is highly imbalanced, with six times more synthetic samples than human samples being considered. The results reported in the table are weighted by class proportion, to account for the aforementioned imbalance.

We repeated these experiments but considering the same proportion of human and synthetic samples for training/validation/testing. Table IV summarizes the results. We can observe that the differences between input representations are more apparent, with four absolute points of difference in terms of AUC, Accuracy, and F-score. These results reject the idea that data imbalance may benefit the velocity-based classifier. Rather, the reason why both classifiers performed similarly in Table III is because they learned much more about the distribution of the synthetic data than the human data, so the nuances and details of real human movements became less noticeable to the classifiers, as they were less frequently represented in the training data.

## VI. DISCUSSION

As shown in Fig. 3, our classifier can effectively distinguish whether a trajectory was synthetically generated, achieving perfect (or very close to 100%) AUC scores across all datasets when using trajectory data as input. Using velocity data as input proved to be more challenging for classification, as we observed lower AUC scores overall, with a minimum value of 65% achieved in the ProjectedSign dataset. A similar observation can be made in Fig. 4, where a lower EER (often 0%) was observed when using trajectory data as input. Given the results of our comprehensive study, we can conclude that it is possible to tell human and machine handwriting apart because, as the saying goes, the devil is in the details.

The research literature has shown that computational models like $\Sigma\Lambda$ can synthesize human movements with remarkable precision. However, replicating intricate movement details related to high-frequency components remains an open research question. It turns out that human handwriting movements are imperfect, in the sense that they contain non-stochastic spatiotemporal variations that computational models are unable to reproduce reliably. These minor, sometimes imperceptible, motor variations are rapidly recognized by our GRU classifier.

To shed more light on our experimental results, we estimated the movement complexity in each dataset. Concretely, we analyzed the distribution of the number of extracted stroke components, denoted as nbLog in the research literature, which is linked to the complexity of handwriting movements [57]. A lower nbLog value, in conjunction with a higher signal-to-noise ratio (SNR), indicates a better adherence to the lognormality principle [58]. But nbLog alone is not comparable across datasets, since complexity is higher if the input signal is longer, so we normalize nbLog by the duration of the input signal, nbLog/s, to make results more comparable.

We use the iDeLog method [59] to detect the number of components in the synthesized specimens, to make it independent from the $\Sigma\Lambda$ synthesizers we used (see Section III-A). iDeLog also allows for greater flexibility in calculating the number of components for handwriting acquired with different digitizers. Our hypothesis relies on the dissimilarities between distributions: the higher the overlap between human and synthetic distributions, the better the synthetic data would approximate the human data.

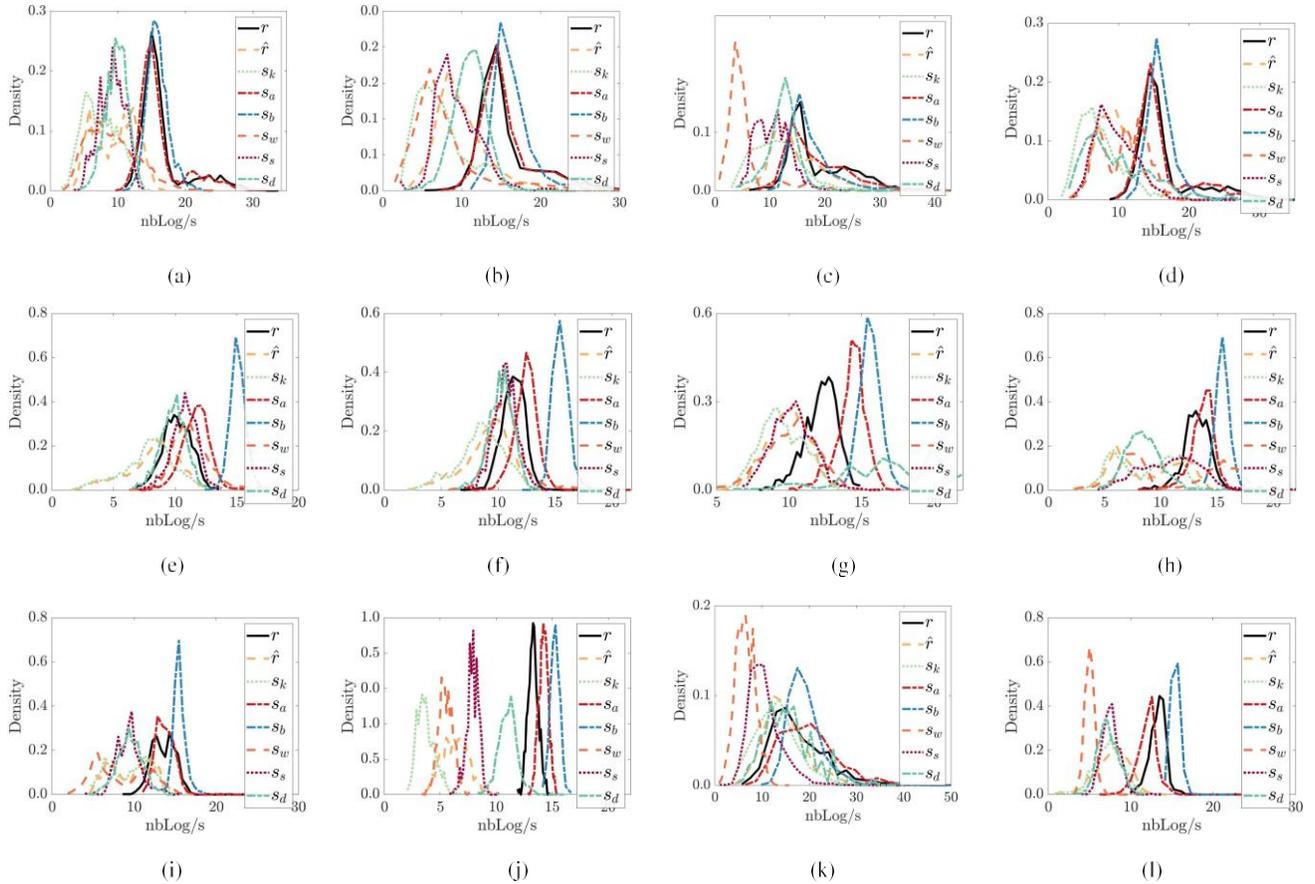

Fig. 9. Density distributions of nbLog/s ($s^{-1}$) analyzed with iDeLog [59] in terms of real samples ($r$), $\Sigma\Lambda$ reconstructed samples ($\hat{r}$), synthesized samples with the kinematic model ($s_k$), affine transformations ($s_a$), BeCAPTCHA ($s_b$), DeepWriteSYN ($s_w$), SDT ($v_s$), and DHG ($s_d$). (a) \$1-GDS. (b) \$N-MMG$_S$. (c) \$N-MMG$_F$. (d) Chars74k. (e) Biosecure. (f) MCYT-100. (g) SUSIGv. (h) BioSign$_S$. (i) BioSign$_F$. (j) Mouse. (k) MobileTouch. (l) ProjectedSign.

In principle we could assume that the smoother (and more predictable) the human movement, the closer its synthetic version will be and therefore the harder it will be to classify. The reason is that a computer-generated movement is more "regular" than a real human movement, as human handwriting is articulated by heterogeneous factors, such as the person's stance, posture, cognitive state, etc. These regularities in artificially generated handwriting are more easily detected while analyzing movement dynamics rather than analyzing their visual appearance [6].

As can be observed in Fig. 9, there are subtle differences between human and synthetic data. Interestingly, we can identify three groups across all datasets. The first group comprises the real and affine-wise data. Their distributions tend to appear centrally in the plots and are very similar in many cases, which in principle indicates that these specimens would be harder to classify. However, as we already know, this is not the case. The second group comprises $\Sigma\Lambda$, DHG, STD, and DeepWrite. Notably, a consistent behavior was observed in the distribution plots. We should point out that these approaches do not operate on a geometric space (as the affine approach does). More interestingly, the $\Sigma\Lambda$ distributions are systematically shifted to the left as compared to the real data. The reason is the signal processing in ScriptStudio, which removes the imperfections from the input signal as a result of its preprocessing, which results in a reduced number of lognormals detected by iDeLog. Finally, the third group consists solely of BeCAPTCHA data. We note that their distributions are systematically shifted to the right as compared to the real data. The reason is that those samples contain more number of components than the samples generated by the other approaches, as can be observed in Fig. 2. Therefore, iDeLog needs more lognormals to approximate the BeCAPTCHA-generated samples, which results in a higher number of nbLog/s values as a result.

Other interesting observations are worth of discussion. First, signatures collected with good-quality input devices, such as those in the Biosecure and MCYT-100 datasets, have lower nbLog/s values. This happens to be correlated with the classification performance of our classifier when using velocity as input representation; cf. Fig. 3. On the other hand, in the Mouse dataset, both the $\Sigma\Lambda$ reconstructions and the kinematic-wise samples have a shifted distribution. This is again caused by the signal processing of ScriptStudio. It is also correlated with the lower performance observed in Fig. 3. This was somehow expected, since this dataset comprises mostly straight movements. Therefore, a few lognormals per second are expected to model such movements. What is interesting, however, is that, obeying the lognormality principle, the number of extracted lognormals can explain the classification performance: the lower the nbLog/s the lower the AUC scores.

Finally, we should point out that our work has a noteworthy limitation. While our classifier is independent of the writer, synthesizer, and language, aiming at a general-purpose reverse Turing test, it requires a collection of examples of human and machine-generated movements to be trained on. Most of our datasets comprise signatures, gestures, and mouse movements. Only two datasets (Chars74k and MobileTouch) comprise handwritten characters, which happen to be in English. Future work should consider handwritten characters in non-Latin writing scripts, such as Chinese or Arabic. Future work could also consider a one-class classification approach, where for example the model is trained on human movements only. This task can be framed as an anomaly detection problem instead of the binary classification problem we have tackled in this work. In addition, future work could also consider explainability tools in order to better understand the discriminating capability of the trained models. Looking forward into the future, it would be interesting to consider other input representations, following emerging research directions on the analysis of high-dimensional inertial data [60], [61], [62].

## VII. Conclusion

We have contributed a comprehensive study and a very accurate classifier to tell human and machine handwriting movements apart. It can distinguish many different types of symbols, from isolated characters and digits, to gestures, pointing traces, and signatures. Our experiments prove the robustness and generalizability of our approach to several data generation methods, to few-shot training, and to out-of-domain classification.

Our results are key not only to biometric systems, but also to other research fields that need to understand the nuanced differences between human and machine movements. For example, in robotics, our model can be used to improve the realism of synthetic but human-like end-effector trajectories, which is a very demanding task for robot arms with flexible links [63]. In sum, our approach not only allows distinguishing next-generation bots trying to impersonate human handwriting, but also has important implications for human behavior analysis and motor control understanding.

## Acknowledgment

The experiments presented in this article were carried out using the HPC facilities of the University of Luxembourg (https://hpc.uni.lu/) and the computing resources provided by the HICUP lab from the University of Primorska. The work of Nuwan T. Attygalle was completed when he was with the University of Primorska, Slovenia.

## References

[1] A. Acien, A. Morales, J. Fierrez, and R. Vera-Rodriguez, "BeCAPTCHA-Mouse: Synthetic mouse trajectories and improved bot detection," *Pattern Recognit.*, vol. 127, Jul. 2022, Art. no. 108643.
[2] S. Shirali-Shahreza and M. Shirali-Shahreza, "Multilingual highlighting CAPTCHA," in *Proc. ITNG*, 2011, pp. 447–452.
[3] L. A. Leiva and F. Álvaro, "µcaptcha: Human interaction proofs tailored to touch-capable devices via math handwriting," *Int. J. Hum.-Comput. Interact.*, vol. 31, no. 7, pp. 457–471, 2012.
[4] M. Diaz-Cabrera, M. A. Ferrer, and A. Morales, "Cognitive inspired model to generate duplicated static signature images," in *Proc. ICFHR*, 2014, pp. 61–66.
[5] L. A. Leiva, D. Martín-Albo, and R. Plamondon, "Gestures à Go Go: Authoring synthetic human-like stroke gestures using the kinematic theory of rapid movements," *ACM Trans. Intell. Syst. Technol.*, vol. 7, no. 2, pp. 1–29, 2016.
[6] L. A. Leiva, M. Diaz, M. A. Ferrer, and R. Plamondon, "Human or machine? It is not what you write, but how you write it," in *Proc. ICPR*, 2021, pp. 2612–2619.
[7] L. Tang, W. Kang, and Y. Fang, "Information divergence-based matching strategy for online signature verification," *IEEE Trans. Inf. Forensics Security*, vol. 13, pp. 861–873, 2017.
[8] P.-M. Bernier, R. Chua, and I. M. Franks, "Is proprioception calibrated during visually guided movements?" *Exp. Brain Res.*, vol. 167, no. 2, pp. 292–296, 2005.
[9] Y. Xin et al., "A survey of liveness detection methods for face biometric systems," *Sensor Rev.*, vol. 37, no. 3, pp. 346–356, 2017.
[10] M. Drahansky, "Liveness detection in biometrics," in *Advanced Biometric Technologies*, G. Chetty and J. Yang, Eds., London, U.K.: IntechOpen, 2011.
[11] Y. Elarian, R. Abdel-Aal, I. Ahmad, M. T. Parvez, and A. Zidouri, "Handwriting synthesis: Classifications and techniques," *Int. J. Doc. Anal. Recognit.*, vol. 17, no. 4, pp. 455–469, 2014.
[12] R. Plamondon, "A kinematic theory of rapid human movements: Part-II. Movement time and control," *Biol. Cybern.*, vol. 72, no. 4, pp. 309–320, 1995.
[13] M. Diaz, A. Fischer, M. A. Ferrer, and R. Plamondon, "Dynamic signature verification system based on one real signature," *IEEE Trans. Cybern.*, vol. 48, no. 1, pp. 228–239, Jan. 2018.
[14] M. A. Ferrer et al., "Static and dynamic synthesis of Bengali and Devanagari signatures," *IEEE Trans. Cybern.*, vol. 48, no. 10, pp. 2896–2907, Oct. 2018.
[15] S. Tang and Z. Lian, "Write like you: Synthesizing your cursive online Chinese handwriting via metric-based meta learning," *Comput. Graph. Forum*, vol. 40, no. 2, pp. 141–151, 2021.
[16] Y. Elarian, I. Ahmad, S. Awaida, W. G. Al-Khatib, and A. Zidouri, "An Arabic handwriting synthesis system," *Pattern Recognit.*, vol. 48, no. 3, pp. 849–861, 2015.
[17] A. Graves, "Generating sequences with recurrent neural networks," 2013, *arXiv:1308.0850*.
[18] B. Zhao, J. Tao, M. Yang, Z. Tian, C. Fan, and Y. Bai, "Deep imitator: Handwriting calligraphy imitation via deep attention networks," *Pattern Recognit.*, vol. 104, Aug. 2020, Art. no. 107080.
[19] D. Ha and D. Eck, "A neural representation of sketch drawings," in *Proc. ICLR*, 2018, pp. 1–20.
[20] E. Aksan, T. Deselaers, A. Tagliasacchi, and O. Hilliges, "Cose: Compositional stroke embedding," in *Proc. NeurIPS*, 2020, pp. 1–12.
[21] R. Tolosana, P. Delgado-Santos, A. Perez-Uribe, R. Vera-Rodriguez, J. Fierrez, and A. Morales, "DeepWriteSYN: On-line handwriting synthesis via deep short-term representations," in *Proc. AAAI*, 2021, pp. 1–9.
[22] E. Aksan, F. Pece, and O. Hilliges, "Deepwriting: Making digital ink editable via deep generative modeling," in *Proc. CHI*, 2018, pp. 1–14.
[23] A. Kotani, S. Tellex, and J. Tompkin, "Generating handwriting via decoupled style descriptors," in *Proc. ECCV*, 2020, pp. 764–780.
[24] A. K. Bhunia, S. Khan, H. Cholakkal, R. M. Anwer, F. S. Khan, and M. Shah, "Handwriting transformers," in *Proc. ICCV*, 2021, pp. 1–9.
[25] G. Dai et al., "Disentangling writer and character styles for handwriting generation," in *Proc. CVPR*, 2023.
[26] T. Luhman and E. Luhman, "Diffusion models for handwriting generation," 2020, *arXiv:2011.06704*.
[27] M.-S. Ren, Y.-M. Zhang, Q.-F. Wang, F. Yin, and C.-L. Liu, "Diff-writer: A diffusion model-based stylized online handwritten Chinese character generator," in *Proc. ICONIP*, 2023, pp. 86–100.
[28] M. Diaz, A. Mendoza-García, M. A. Ferrer, and R. Sabourin, "A survey of handwriting synthesis from 2019 to 2024: A comprehensive review," *Pattern Recognit.*, vol. 162, Jun. 2025, Art. no. 111357.
[29] C. O'Reilly and R. Plamondon, "Development of a Sigma–Lognormal representation for on-line signatures," *Pattern Recognit.*, vol. 42, no. 12, pp. 3324–3337, 2009.
[30] S. Lai, L. Jin, Y. Zhu, Z. Li, and L. Lin, "SynSig2Vec: Forgery-free learning of dynamic signature representations by sigma lognormal-based synthesis and 1D CNN," *IEEE Trans. Pattern Anal. Mach. Intell.*, vol. 44, no. 10, pp. 6472–6485, Oct. 2022.

[31] T. L. Kumar and L. A. Leiva, "Attentive sequence-to-sequence modeling of stroke gestures articulation performance," *IEEE Trans. Hum. Mach. Syst.*, vol. 51, no. 6, pp. 663–672, Dec. 2021.

[32] K. He, X. Zhang, S. Ren, and J. Sun, "Deep residual learning for image recognition," in *Proc. CVPR*, 2016, pp. 1–9.

[33] M. Sandler, A. Howard, M. Zhu, A. Zhmoginov, and L.-C. Chen, "MobileNetV2: Inverted residuals and linear bottlenecks," in *Proc. CVPR*, 2018, pp. 1–11.

[34] O. Ronneberger, P. Fischer, and T. Brox, "U-Net: Convolutional networks for biomedical image segmentation," in *Proc. MICCAI*, 2015, pp. 234–241.

[35] K. Cho, B. van Merrienboer, C. Gulcehre, F. Bougares, H. Schwenk, and Y. Bengio, "Learning phrase representations using RNN encoder-decoder for statistical machine translation," in *Proc. EMNLP*, 2014, pp. 1724–1734.

[36] S. Hochreiter and J. Schmidhuber, "Long short-term memory," *Neural Comput.*, vol. 9, no. 8, pp. 1735–1780, Nov. 1997.

[37] R. Jozefowicz, W. Zaremba, and I. Sutskever, "An empirical exploration of recurrent network architectures," in *Proc. ICML*, 2015, pp. 1–9.

[38] J. Chung, C. Gulcehre, K. Cho, and Y. Bengio, "Empirical evaluation of gated recurrent neural networks on sequence modeling," in *Proc. NeurIPS Workshops*, 2014, pp. 1–9.

[39] R. Pascanu, T. Mikolov, and Y. Bengio, "On the difficulty of training recurrent neural networks," in *Proc. ICML*, 2013, pp. 1–9.

[40] Y. Bengio, P. Frasconi, and P. Simard, "The problem of learning long-term dependencies in recurrent networks," in *Proc. IJCNN*, 1993, pp. 1183–1188.

[41] Y. Bengio, P. Simard, and P. Frasconi, "Learning long-term dependencies with gradient descent is difficult," *IEEE Trans. Neural Netw.*, vol. 5, no. 2, pp. 157–166, Mar. 1994.

[42] G. Stragapede, R. Vera-Rodriguez, R. Tolosana, and A. Morales, "BehavePassDB: Public database for mobile behavioral biometrics and benchmark evaluation," *Pattern Recognit.*, vol. 134, Feb. 2023, Art. no. 109089.

[43] M. Liwicki et al., "SigComp11: Signature verification competition for onand offline skilled forgeries," in *Proc. ICDAR*, 2011, pp. 1480–1484.

[44] P. Zhang, J. Jiang, Y. Liu, and L. Jin, "MSDS: A large-scale Chinese signature and token digit string dataset for handwriting verification," in *Proc. NeurIPS*, vol. 35, 2022, pp. 1–13.

[45] J. O. Wobbrock, A. D. Wilson, and Y. Li, "Gestures without libraries, toolkits or training: A $1 recognizer for user interface prototypes," in *Proc. UIST*, 2007, pp. 159–168.

[46] L. Anthony and J. O. Wobbrock, "A lightweight multistroke recognizer for user interface prototypes," in *Proc. GI*, 2010, pp. 245–252.

[47] T. E. de Campos, B. R. Babu, and M. Varma, "Character recognition in natural images," in *Proc. VISAPP*, 2009, pp. 273–280.

[48] A. Kholmatov and B. Yanikoglu, "SUSIG: An on-line signature database, associated protocols and benchmark results," *Pattern Anal. Appl.*, vol. 12, no. 3, pp. 227–236, 2009.

[49] J. Galbally, M. Diaz-Cabrera, M. A. Ferrer, M. Gomez-Barrero, A. Morales, and J. Fierrez, "On-line signature recognition through the combination of real dynamic data and synthetically generated static data," *Pattern Recognit.*, vol. 48, no. 9, pp. 2921–2934, 2015.

[50] J. Ortega-Garcia et al., "MCYT baseline corpus: A bimodal biometric database," *IEEE Proc. Vis. Image Signal Process.*, vol. 150, no. 6, pp. 170–180, 2003.

[51] R. Tolosana, J. Gismero-Trujillo, R. Vera-Rodriguez, J. Fierrez, and J. Ortega-Garcia, "MobileTouchDB: Mobile touch character database in the wild and biometric benchmark," in *Proc. CVPR Workshops*, 2019, pp. 2306–2314.

[52] R. Tolosana, R. Vera-Rodriguez, J. Fierrez, A. Morales, and J. Ortega-Garcia, "Benchmarking desktop and mobile handwriting across COTS devices: The e-BioSign biometric database," *PLoS ONE*, vol. 12, no. 5, 2017, Art. no. e0176792.

[53] C. Shen, Z. Cai, X. Guan, and R. Maxion, "Performance evaluation of anomaly-detection algorithms for mouse dynamics," *Comput. Secur.*, vol. 45, pp. 156–171, Sep. 2014.

[54] S. K. Behera, D. P. Dogra, and P. P. Roy, "Analysis of 3D signatures recorded using leap motion sensor," *Multimed. Tools Appl.*, vol. 77, no. 11, pp. 14029–14054, 2018.

[55] E. Griechisch, J. R. Ward, and G. Hanczár, "Anomalies in measuring speed and other dynamic properties with touchscreens and tablets," in *Proc. BIOSIG*, 2019, pp. 1–6.

[56] M. Tan et al., "Out-of-domain detection for low-resource text classification tasks," in *Proc. EMNLP-IJCNLP*, 2019, pp. 3566–3572.

[57] R. Vera-Rodriguez, R. Tolosana, J. Hernandez-Ortega, A. Morales, J. Fierrez, and J. Ortega-Garcia, "Modeling the complexity of biomechanical tasks using the lognormality principle: Applications to signature recognition and touch-screen children detection," in *Proc. ICPRAI*, 2018, pp. 1–6.

[58] R. Plamondon, A. Marcelli, M. A. Ferrer, and R. Plamondon, Eds., *The Lognormality Principle and Its Applications in e-Security, e-Learning and e-Health*. Singapore: World Sci., 2020.

[59] M. A. Ferrer, M. Diaz, C. Carmona-Duarte, and R. Plamondon, "iDeLog: Iterative dual spatial and kinematic extraction of sigma-lognormal parameters," *IEEE Trans. Pattern Anal. Mach. Intell.*, vol. 42, no. 1, pp. 114–125, Jan. 2020.

[60] W.-C. Lai and H. Schröter, "Ubicomp digital 2020–handwriting classification using a convolutional recurrent network," 2020, *arXiv:2008.01078*.

[61] F. Imbert, E. Anquetil, Y. Soullard, and R. Tavenard, "Mixture-of-experts for handwriting trajectory reconstruction from IMU sensors," *Pattern Recognit.*, vol. 161, May 2025, Art. no. 111231.

[62] Y. Wang and Y. Zhao, "Chinese inertial GAN for handwriting signal generation and recognition," 2025, *ChinaXiv:202506.00010*.

[63] A. D. Luca, L. Lanari, and G. Ulivi, "End-effector trajectory tracking in flexible arms: Comparison of approaches based on regulation theory," in *Advanced Robot Control*, vol. 162, C. de Wit, Ed., Berlin, Germany: Springer, 1991, doi: 10.1007/BFb0039272.

**Luis A. Leiva** received the first B.Sc. degree in industrial design, the second B.Sc. degree in industrial engineering, the M.Sc. degree in electrical engineering, the M.A.S. degree in pattern recognition and artificial intelligence, and the Ph.D. degree in computer science from the Universitat Politècnica de València, Valencia, Spain, in 2001, 2005, 2007, 2008, and 2012, respectively.

He is an Associate Professor with the Computer Science Department, University of Luxembourg, Esch-sur-Alzette, Luxembourg, where he leads the Computational Interaction group. His research interests lie at the intersection of human–computer interaction and machine learning.

**Moises Diaz** received the M.Tech. and M.Sc. degrees in engineering, the M.Ed. degree in secondary education, and the Ph.D. degree in engineering, and from the La Universidad de Las Palmas de Gran Canaria, Las Palmas, Spain, in 2010, 2011, 2013, and 2016, respectively.

He is an Associate Professor with the University of Las Palmas de Gran Canaria. His research interests include pattern recognition, document analysis, handwriting recognition, biometrics, and intelligent transportation systems.

**Nuwan T. Attygalle** received the B.Sc. degree in physical science from the University of Sri Jayewardenepura, Nugegoda, Sri Lanka, in 2017, and the Ph.D. degree in computer science from the University of Primorska, Koper, Slovenia, in 2024.

He is currently a Postdoctoral Scholar with the Université Catholique de Louvain, Louvain-la-Neuve, Belgium. His research interests include interaction techniques, gesture recognition, machine learning, and artificial intelligence.

**Miguel A. Ferrer** received the M.Sc. and Ph.D. degrees in electrical engineering from the Universidad Politécnica de Madrid, Madrid, Spain, in 1988 and 1994, respectively.

He joined the University of Las Palmas de Gran Canaria, Las Palmas, Spain, in 1989, where he is a Full Professor. He established the Digital Signal Processing Research Group in 1990. His current research interests include pattern recognition, biometrics, audio quality, and computer vision applications to fisheries and aquaculture.

**Réjean Plamondon** (Life Fellow, IEEE) received the B.Sc. degree in physics and the M.Sc. and Ph.D. degrees in electrical engineering from the Université Laval, Québec, QC, Canada, in 1973, 1975, and 1978, respectively.

In 1978, he joined the Faculty of École Polytechnique, Université de Montréal, Montréal, QC, Canada, as a Full Professor, where he is also the Head of the Laboratoire Scribens.

Prof. Plamondon received the IAPR/ICDAR outstanding achievement award in 2013, for his contributions to human movement understanding and its applications.